\documentclass[lettersize,journal]{IEEEtran}
\usepackage{cite}
\usepackage{amsmath,amssymb,amsfonts}
\usepackage{algorithmic}
\usepackage{graphicx}
\usepackage{textcomp}
\usepackage{xcolor}
\usepackage{stfloats}
\usepackage{booktabs}
\usepackage{multirow}
\usepackage{array}
\usepackage{tikz}
\usetikzlibrary{arrows.meta, positioning, shapes.geometric}
\def\BibTeX{{\rm B\kern-.05em{\sc i\kern-.025em b}\kern-.08em
    T\kern-.1667em\lower.7ex\hbox{E}\kern-.125emX}}

\begin{document}

\title{Enhancing Science Classroom Discourse Analysis through Joint Multi-Task Learning for Reasoning-Component Classification}

\author{Jiho Noh,
        Mukhesh Raghava Katragadda,
        Raymond Carl,
        and Soon Lee%
\thanks{J. Noh, M. R. Katragadda, and R. Carl are with the Department of
Computer Science, Kennesaw State University, Kennesaw, GA 30144 USA.
E-mail: jnoh3@kennesaw.edu; mkatrag1@students.kennesaw.edu;
rcarl3@students.kennesaw.edu.}%
\thanks{S. Lee is with the Bagwell College of Education, Kennesaw State
University, Kennesaw, GA 30144 USA. E-mail: slee263@kennesaw.edu.}}

\markboth{Preprint}%
{Noh \MakeLowercase{\textit{et al.}}: Enhancing Science Classroom Discourse Analysis}

\maketitle

\begin{abstract}
Analyzing the reasoning patterns of students in science classrooms is critical
for understanding knowledge construction mechanism and improving instructional
practice to maximize cognitive engagement, yet manual coding of classroom
discourse at scale remains prohibitively labor-intensive. We present an
automated discourse analysis system (ADAS) that jointly classifies teacher and
student utterances along two complementary dimensions: \emph{Utterance Type}
and \emph{Reasoning Component} derived from our prior CDAT framework. To
address severe label imbalance among minority classes, we (1) stratify-resplit
the annotated corpus, (2) apply LLM-based synthetic data augmentation targeting
minority classes, and (3) train a dual-probe head RoBERTa-base classifier. A
zero-shot GPT-5.4 baseline achieves macro-F1 of 0.582 on UT and 0.412 on RC;
fine-tuned RoBERTa with augmentation surpasses it on the 10-class UT task
(0.635) but not on the 4-class RC task, where GPT-5.4 remains the strongest
system. Beyond classification, we conduct discourse pattern analyses
including UT$\times$RC co-occurrence profiling, Cognitive Complexity Index
(CCI) computation per session, lag-sequential analysis, and IRF chain analysis
revealing that teacher \emph{Feedback-with-Question} (Fq) moves are the most
consistent antecedents of student inferential reasoning (SR-I). Our results
demonstrate that LLM-based augmentation meaningfully improves UT minority-class
recognition, and that the structural simplicity of the RC task makes it
tractable even for lexical baselines.
\end{abstract}


\section{Introduction}

\IEEEPARstart{U}{nderstanding} how students reason during classroom instruction is one of the
primary challenges in educational research. Classroom discourse --- the
structured exchange of ideas between teachers and students --- serves as the
primary mechanism through which knowledge is co-constructed
\cite{Chin2006-yb,Mercer1999-up}. Prior research on classroom
discourse analysis remains predominantly qualitative and limited to
surface-level features such as turn-taking frequency and basic statistics of
dialogue acts, leaving the richer cognitive dimension of \emph{what kind of
reasoning} is being expressed largely unmeasured at scale. This gap limits our
ability to systematically characterize the quality of classroom interactions
and to identify instructional moves that effectively promote higher-order
scientific reasoning.

Discourse analysis frameworks have a long history in education research
\cite{Rogers2005-cv, Forman1995-rf}. Flanders' Interaction Analysis Categories
(FIAC) \cite{Flanders1970-ip,Odiri-Amatari2015-kh} and the
Initiation-Response-Feedback (IRF) model of Sinclair and Coulthard
\cite{Sinclair1975-ij} established the foundational vocabulary for describing
teacher--student interaction. These frameworks capture the \emph{structural}
patterns of classroom talk --- who speaks, when, and in response to what ---
but do not systematically characterize the \emph{cognitive quality} of what is
being said. Our prior work, Classroom Discourse Analysis Tool (CDAT) framework,
addresses this gap by positioning classroom utterances along a continuum from
everyday (inductive) to scientific (deductive) reasoning, providing both an
utterance-type taxonomy and a reasoning-component annotation scheme
\cite{Lee2018-hw}. 

A prevalent challenge in educational discourse analysis is the labor-intensive
nature of manual transcription and coding. The data collection process
generally results in a substantial volume of raw audio files (or text files
automatically generated via speech-to-text tools), but only a small fraction of
these are manually annotated with discourse codes due to the time and expertise
required. This creates a bottleneck for scaling up analyses to larger datasets,
which is necessary for robust pattern discovery and generalizable insights,
particularly in the context of machine learning-based classification models
that require substantial labeled data for training to attain satisfactory
performance.

Specifically, the utterance and reasoning type classification in accordance
with our Automated Discourse Analysis System (ADAS) framework presents several
unique challenges. There exists a high degree of subjectivity among human
coders when labeling certain types of utterance, coupled with significant class
imbalance in the data, which poses obstacles to achieving reliable automated
coding. Additionally, the scarcity of large, diverse, and consistently labeled
training datasets further hampers the development of robust classification
models.

The objective of this study is to develop and evaluate an automated discourse
analysis system (ADAS) that jointly classifies teacher and student utterances
along two complementary dimensions --- Utterance Type (UT) and Reasoning
Component (RC) --- using a pre-trained transformer-based model trained on a
labeled corpus of science classroom transcripts. To address the severe class
imbalance that characterizes naturally occurring classroom data, we combine a
principled 4-class revision of the original CDAT coding scheme with LLM-based
synthetic data augmentation and focal-loss training. Beyond model development,
this study pursues a deeper analytical goal: to characterize the
\emph{patterns} of classroom discourse at the level of cognitive engagement,
including how reasoning-component distributions evolve within a lesson and
which instructional moves reliably precede higher-order student reasoning.

Accordingly, we aim to reveal not only \emph{what} kinds of utterances populate
science classroom discourse, but \emph{how} the cognitive quality of student
contributions varies across instructional contexts and over the course of a
lesson. By combining automated classification with discourse pattern analysis ---
co-occurrence profiling, lag-sequential teacher-move analysis, and Cognitive
Complexity Index (CCI) time series --- we seek to provide both a scalable
coding tool and actionable insights for instructional design.

Our principal contributions are:
\begin{itemize}
  \item A revised RC coding scheme that collapses the original CDAT 6-class
    scheme along theoretically motivated boundaries, improving learnability
    without sacrificing framework traceability.
  \item A cross-conditioned dual-head RoBERTa-base model that jointly classifies
    UT and RC with symmetric task conditioning.
  \item A comprehensive discourse pattern analysis including session-level CCI
    time series and lag-sequential teacher-move to student-reasoning analysis.
  \item A human vs. pseudo-label comparison of discourse-level findings,
    revealing positional annotation bias as the likely source of the
    end-of-lesson CCI rebound in human annotations.
\end{itemize}

\section{Literature Review}

\subsection{Conceptual Framework and Coding Scheme for Science Classroom Discourse Analysis}

The IRF triplet --- teacher Initiation, student Response, teacher Feedback ---
remains the foundational structural unit of classroom interaction
\cite{Sinclair1975-ij,Mehan2014-jf}. While IRF analysis reveals participation
structure, it is agnostic to the cognitive level of contributions. Accountable
Talk \cite{Michaels2008-ys} and dialogic teaching frameworks
\cite{Alexander2008-xs} extend IRF by characterizing the quality of classroom
talk, which motivated our study's focus on reasoning components; we can relate
``rigorous thinking'' in Accountable Talk to our reasoning component and
``Purposeful'' cumulative reasoning in dialogic teaching to our Cognitive
Complexity Index. 

In science education specifically, the CDAT framework \cite{Lee2018-hw}
provides a fine-grained coding system grounded in the distinction between
everyday and scientific reasoning. The framework aligns with Next Generation
Science Standards (NGSS) \cite{NGSS-Lead-States2013-ed} and Bloom's taxonomy
\cite{Bloom1956-ni}, providing a theoretically coherent hierarchy from informal
experiential knowledge (EX) through declarative scientific knowledge (SK) and
observational data (OD), to pattern identification (PD) and model/theory
construction (MT). Table~\ref{tab:ut_labels} and Table~\ref{tab:rc_labels}
summarize the coding schemes used in the framework. 

\begin{table}[t]
\centering
\caption{Utterance Type (UT) coding scheme (10 classes).}
\label{tab:ut_labels}
\begin{tabular}{@{}llp{3.7cm}@{}}
\toprule
\textbf{Code (Name)} & \textbf{Speaker} & \textbf{Description} \\
\midrule
Q (Question)             & Teacher & Inquiry posed to students \\
P (Prompt)               & Teacher & Cue to guide student thinking \\
E (Explanation/Example)  & Teacher & Detailed information or illustration \\
Fy (Feedback—Confirm.)    & Teacher & Affirm / repeat student response \\
Fs (Feedback—Statement)   & Teacher & Feedback with additional content \\
Fq (Feedback—Question)    & Teacher & Feedback that probes further \\
Rs (Response—Statement)   & Student & Substantive declarative answer \\
Rq (Response—Question)    & Student & Student inquiry or rephrased Q \\
Ry (Response—Confirm.)    & Student & Agreement or minimal reply \\
O (Other)                & Both    & Off-topic, inaudible, procedural \\
\bottomrule
\end{tabular}
\end{table}

\begin{table}[t]
\centering
\caption{Reasoning Component (RC) coding scheme (6 classes).}
\label{tab:rc_labels}
\begin{tabular}{@{}lp{5cm}@{}}
\toprule
\textbf{Code (Name)} & \textbf{Description} \\
\midrule
EX (Experience)
  & Everyday experiences or teacher anecdotes\\
SK (Scientific Knowledge)
  & Scientific concepts and definitions\\
OD (Observation/Data)
  & Descriptions of natural phenomena or quantified value\\
PD (Patterns from Data)
  & Data-driven reasoning and generalization\\
MT (Models/Theories)
  & Explanations from observations and data patterns\\
O (Other)
  & No reasoning content to the instructional topic \\
\bottomrule
\end{tabular}
\end{table}

\subsection{Challenges in Sequential Sentence Classification}

The classification of utterance types within a classroom discourse context can
be framed as a sequential sentence classification
task~\cite{Yogatama2017-wh,Cohan2019-yz}. The task is designed as
``sequential'' to differentiate it from the more general sentence (or document)
classification task, in which sentences are classified independently, without
utilizing local context. Context is particularly crucial in classroom
discourse, as the semantic meaning and functional role of an utterance can be
significantly influenced by surrounding utterances.

Prior approaches to this task have employed various neural network
architectures, often based on hierarchical models that combine
transformer-based encoders with recurrent layers such as BiLSTM, and
CRF~\cite{Lafferty2001-bb} for sequence
labeling~\cite{Jin2018-it,Chang2019-eq}. These models aim to capture both the
rich semantic representations provided by pre-trained language models and the
contextual dependencies among utterances. Nevertheless, due to the inherent
characteristics of conversational data, certain limitations remain, such as
static dependency modeling, which may not adequately capture the nuanced
dynamics of discourse.

Recent advancements in large language models (LLMs) have introduced new
possibilities for sequential text classification. LLMs can be employed in
various stages of the classification pipeline, including feature extraction,
fine-tuning, transfer learning, and direct inference through
prompting~\cite{Lan2024-ae,Sun2024-su}. The use of LLMs for text classification
tasks, however, still faces challenges such as the difficulty in training on
limited labeled data, potential privacy concerns, and the need for careful
prompt engineering to elicit accurate responses.

\subsection{Strategies for Mitigating the Issue of Insufficient Labeled Data}

A prevalent challenge in developing machine learning models with real-world
data in a specific domain is the lack of sufficiently labeled data, which can
lead to suboptimal model performance (e.g., overfitting, poor
generalization)~\cite{Li2023-fg}. Due to the labor-intensive nature of manual
annotation, particularly within specialized domains, acquiring a large and
diverse labeled dataset can be challenging. Common strategies to address this
issue include data augmentation techniques and semi-supervised learning (SSL)
methods: Specifically, in the context of text classification tasks, two widely
adopted approaches are the use of LLM-generated synthetic data and the
application of pseudo-labeling techniques~\cite{Duarte2023-ia,Li2024-ul}.

With the advent of LLMs, data augmentation has become more feasible through the
generation of synthetic training examples. With available annotated text data,
LLMs can be prompted to rephrase or generate variations of existing examples,
thereby expanding the training dataset and improving model robustness. However,
text data generated by LLM may potentially include biased content, which cannot
be comprehensively evaluated either automatically or by human
assessment~\cite{Zheng2023-jh}. Additionally, a general-domain LLM may generate
inaccurate augmentation results due to its deficiency in domain-specific
knowledge~\cite{Dai2025-to}. Still, LLM-based data augmentation has shown
promise in enhancing model performance across various NLP tasks.

Another effective approach is semi-supervised learning, which utilizes both
labeled and unlabeled data during the training
process~\cite{Lee2013-xo,Zhang2021-jw,Chen2022-nm}, wherein a model trained on
labeled data can be used to predict labels for unannotated data. This
pseudo-labeled data is than incorporated into the training set, effectively
increasing the amount of labeled data available for model training. One of the
benefits of SSL is that it can leverage the vast amount of unlabeled data
containing real examples from data collection. A commonly revealed challenge
with self-training is, however, the bias in the pseudo-labels, which can lead
to error accumulation during the iterative self-training
process~\cite{Chen2022-nm}. Various strategies have been proposed to mitigate
this issue, such as confidence thresholding, where only predictions with high
confidence are added to the training set, and consistency regularization, which
encourages the model to produce consistent predictions under different
perturbations of the input
data~\cite{Chen2022-nm,Wang2022-jz,Radhakrishnan2023-xd}.

\subsection{Approaches for Mitigating Class Imbalance}

Classroom discourse datasets also have class imbalance. Common utterance types
(e.g., Rs) occur with significantly greater frequency compared to rare
utterance types (e.g., Fy). This imbalance makes models favorable to major
classes and perform poorly on rare classes~\cite{Taskiran2025-rq}. Two
approaches can handle imbalanced data: resampling techniques and algorithmic
approaches. Resampling techniques oversample rare classes or downsample major
classes to balance the distribution~\cite{Taskiran2025-rq}. Algorithmic
approaches adjust the loss function to give more weight to minority class
errors. Cost-sensitive learning assigns higher loss weights to rare class
errors during training~\cite{Elkan2001-un}.

\section{Methodology}

\subsection{Data}
\label{sec:data}

In this study, we utilize a dataset comprising nine lecture transcripts of 
science classes collected from two middle school and one high school teachers in
the United States. All data, including video/audio recordings and transcripts,
were collected with informed consent and Institutional Review Board (IRB)
approval, ensuring appropriate consent for secondary research use.

From the audio/video recordings, we generated transcripts using an automatic
speech recognition (ASR) tool, followed by speaker role identification and
semantic segmentation into individual utterances. Collectively, the dataset
contains 1,782 manually labeled utterances, each annotated according to the
ADAS framework described in the early section.

\begin{figure}[!t]
  \centering
  \includegraphics[width=0.95\linewidth]{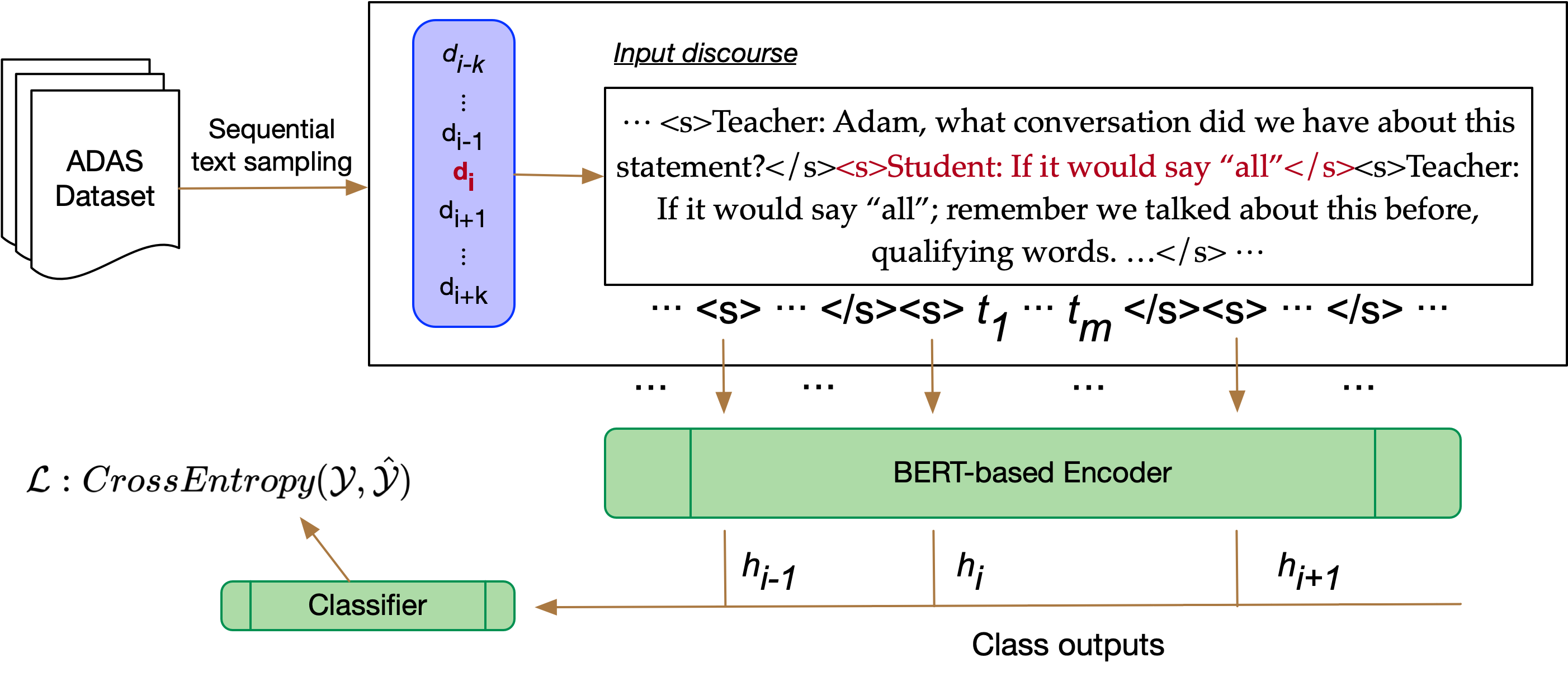}
  \caption{Structure of RoBERTa-based contextualized encoder for label classification}
  \label{fig:model-architecture}
\end{figure}

Each utterance is represented as a context window of the $k$ preceding and $k$
following utterances, each prepended with a speaker tag
(\texttt{Teacher/Student}: or its corresponding embeddings). The input format
follows a structured tokenization scheme (Figure~\ref{fig:model-architecture})
where $d_i$ is the target utterance and $d_{i \pm n}$ are context utterances.
Context windows are pre-computed per conversation before splitting to prevent
cross-split contamination.

\subsubsection{Revised Reasoning Component Coding Scheme}

The original CDAT RC taxonomy defines six classes (Table~\ref{tab:rc_labels}).
Three critical challenges prevent effective automated classification with this
scheme: (1)~extreme label imbalance (MT: 4 examples; EX: 58 examples across
1,782 annotated rows); (2)~non-representative splits drawn from different
teachers and sessions; and (3)~boundary ambiguity between adjacent categories
(SK vs.\ OD; PD vs.\ MT) that depresses inter-annotator agreement. Hence, we
collapse the six-class scheme into a four-class scheme by merging theoretically
adjacent categories along the CDAT framework's own inductive/deductive divide:

\begin{itemize}
  \item \textbf{ER (Everyday Reasoning):} Utterances grounded in informal,
    everyday knowledge --- personal experiences, anecdotes, analogies, or naive
    explanations not derived from systematic scientific inquiry. Subsumes EX.
  \item \textbf{SR-D (Scientific Reasoning---Descriptive):} Utterances that
    state or recall scientific content descriptively --- established concepts,
    definitions, terminology, or reported observations/measurements --- without
    drawing inferences beyond what is directly given. Merges SK and OD.
  \item \textbf{SR-I (Scientific Reasoning---Inferential):} Utterances that
    go beyond description to reason actively with data: identifying patterns,
    making comparisons, generalizing, predicting, constructing explanations, or
    proposing models. Merges PD and MT.
  \item \textbf{O (Others):} Procedural talk, classroom management,
    inaudible segments, and off-topic remarks. Unchanged from original.
\end{itemize}

The mapping from original to revised codes is deterministic
(Table~\ref{tab:rc_remap}) and requires no re-annotation. We devised the
Cognitive Complexity Index (CCI) as a scalar measure of reasoning quality that
assigns values to the revised RC classes: O=0, ER=1, SR-D=2, SR-I=3.

\begin{table}[t]
\centering
\caption{Deterministic mapping from original 6-class CDAT RC scheme to the
revised 4-class scheme adopted in this work.}
\label{tab:rc_remap}
\begin{tabular}{lll}
\toprule
\textbf{Original (6-class)} & \textbf{Revised (4-class)} & \textbf{Rationale} \\
\midrule
EX (Experience)         & ER   & Direct 1:1 \\
SK (Scientific Knowl.)  & SR-D & Both are descriptive acts \\
OD (Obs./Data)          & SR-D & Both are descriptive acts \\
PD (Pattern from Data)  & SR-I & Both are inferential acts \\
MT (Models/Theories)    & SR-I & Both are inferential acts \\
O                       & O   & Unchanged \\
\bottomrule
\end{tabular}
\end{table}

\subsubsection{Class Distribution and Imbalance}

The RC distribution reveals severe imbalance: SR-D dominates at approximately
58\% of labeled examples, while SR-I ($\approx$7.8\%) and ER ($\approx$3.1\%)
are greatly underrepresented. A further challenge arises from the
contextualized input format: each example encodes not only the target utterance
but a window of $k$ surrounding turns. Row-level
stratified splitting --- in which individual utterances are assigned
independently to splits --- therefore can introduce \textbf{data leakage}: context
turns belonging to a training target may simultaneously appear as held-out
targets in validation or test, and vice versa. 

To eliminate this leakage we adopt \textbf{session-level splitting}, assigning
whole sessions to each split so that no session's utterances appear in more
than one partition. With nine sessions available, we enumerate all feasible
6/2/1 train/val/test assignments (252 combinations) and select the partition
that minimizes total Jensen--Shannon (JS) divergence between each split's RC
distribution and the corpus-level distribution, motivated by the work on ``the
Stratification of Multi-Label Data'' by Sechidis et al.~\cite{Sechidis2011-hk}.

The exhaustive search yields JS divergence $= 0.257$, substantially lower than
iterative stratification ($0.519$). The residual divergence is irreducible:
only three sessions contain meaningful SR-I content, making perfect balance
across three splits impossible. Table~\ref{tab:class_distribution} reports
label counts across splits for both tasks.

\begin{table}[hbt]
\centering
\caption{Class distribution by split (session-level 6/2/1 exhaustive-search
  partition; $N=1{,}782$). Session-level assignment prevents context-window
  data leakage. Subtotal \% is over the full corpus.}
\label{tab:class_distribution}
\setlength{\tabcolsep}{4pt}
\begin{tabular}{@{}llrrrr@{}}
\toprule
\textbf{Cat} & \textbf{Label} & \textbf{Train} & \textbf{Val} & \textbf{Test}
  & \textbf{Subtotal (\%)} \\
\midrule
\multirow{10}{*}{UT}
 & Rs  & 237 & 138 &  62 &  437 (24.5\%) \\
 & Fq  &  99 &  85 &  25 &  209 (11.7\%) \\
 & Q   & 131 &  34 &  32 &  197 (11.1\%) \\
 & NA  & 104 &  39 &  50 &  193 (10.8\%) \\
 & Fs  &  90 &  48 &  55 &  193 (10.8\%) \\
 & P   &  81 &  55 &  18 &  154 ( 8.6\%) \\
 & Ry  &  51 &  37 &  16 &  104 ( 5.8\%) \\
 & E   &  80 &   8 &  14 &  102 ( 5.7\%) \\
 & Rq  &  32 &  38 &  28 &   98 ( 5.5\%) \\
 & Fy  &  80 &  10 &   5 &   95 ( 5.3\%) \\
\midrule
\multirow{4}{*}{RC}
 & SR-D & 596 & 265 & 176 & 1{,}037 (58.2\%) \\
 & NA   & 256 & 182 & 112 &   550 (30.9\%) \\
 & SR-I &  97 &  31 &  11 &   139 ( 7.8\%) \\
 & ER   &  36 &  14 &   6 &    56 ( 3.1\%) \\
\midrule
\multicolumn{2}{@{}l}{\textbf{Grand Total}} &
  \textbf{985} & \textbf{492} & \textbf{305} &
  \textbf{1{,}782 (100.0\%)} \\
\bottomrule
\end{tabular}
\end{table}

\subsection{LLM-Based Data Augmentation}

To improve both class balance and overall surface-form diversity, we use GPT-4.1
to generate synthetic variations of existing labeled training utterances. Each
prompt supplies the full UT and RC label taxonomy with definitions, the
surrounding context window (preceding and following utterances with speaker
tags), and the target utterance with its UT and RC labels. The model is
instructed to produce $N$ distinct variations, preserving the original discourse labels,
speaker role, and conversational coherence. The full prompt template is provided
in Appendix~\ref{app:prompt}.

Augmentation proceeds in two optional passes applied to the training split only:

\begin{description}
  \item[Pass 1] --- \textbf{minority-class boost (optional).} \\
    When \texttt{--boost\_minority} is set, utterances whose UT \emph{or} RC class
    count falls below a target ratio $\rho$ of the majority class count are
    identified as boost candidates. For each such utterance the number of extra
    variations generated is $\lceil(\rho \cdot N_{\max} - N_c) / N_c \rceil$,
    where $N_c$ is the current class count and $N_{\max}$ is the majority class
    count.

  \item[Pass 2] --- \textbf{main pass.} \\
    All training utterances receive $n$ variations regardless of class
    membership, expanding the training set by a factor of $(1+n)$ uniformly and
    increasing surface-form diversity across all UT and RC classes.
\end{description}

Each synthetic example is stored with the context and the same UT and RC labels
as the original, allowing us to experiment with different context window sizes.
Table~\ref{tab:aug_var} provides illustrative examples of generated augmented
utterances for a single original utterance.

\begin{table}[hb!]
  \centering
  \small
  \caption{Illustrative examples of generated augmented utterances}
  \label{tab:aug_var}
  \begin{tabular}{@{}lp{6cm}@{}}
    \toprule
    Variations & Utterance \\
    \midrule
    Original & ``\textit{What is the process of photosynthesis?}'' --- role: Teacher, labels: Q/SR-D \\
    Variation 1 & ``\textit{Can someone explain how photosynthesis works?}'' \\
    Variation 2 & ``\textit{Who can describe the steps involved in photosynthesis?}'' \\
    \bottomrule
  \end{tabular}
\end{table}

\subsection{Model Architecture}

\begin{figure}[t]
\centering
\begin{tikzpicture}[
  box/.style={draw, rounded corners, minimum width=2.1cm, minimum height=0.65cm,
              font=\small, align=center},
  arr/.style={-{Stealth[scale=0.8]}, thick}
]
  \node[box, fill=blue!15, minimum width=5.4cm] (roberta) at (0, 0)
        {RoBERTa-base encoder};

  \node[font=\footnotesize] (lctx) at (-2.3, 1.25)
        {$\langle s\rangle d_{t{-}k}\langle/s\rangle\!\cdots$};
  \node[draw, rounded corners=2pt, fill=yellow!30,
        font=\footnotesize, inner sep=4pt] (tgt) at (0, 1.25)
        {$\langle s\rangle\;d_t\;\langle/s\rangle$};
  \node[font=\footnotesize] (rctx) at (2.3, 1.25)
        {$\cdots\!\langle s\rangle d_{t{+}k}\langle/s\rangle$};
  \node[font=\scriptsize, gray, align=center] at (0, 0.62)
        {$\underbrace{\hspace{5.4cm}}_{2k+1\text{ utterances, each wrapped in }
          \langle s\rangle\cdots\langle/s\rangle}$};

  \node[box, fill=orange!25] (bos) at (-1.8, 2.55)
        {$h_{\langle s\rangle}^{(t)}$\\{\footnotesize BOS probe}};
  \node[box, fill=green!25]  (eos) at ( 1.8, 2.55)
        {$h_{\langle/s\rangle}^{(t)}$\\{\footnotesize EOS probe}};

  \node[box, fill=orange!35] (utp) at (-1.8, 3.7) {UT proj.\\$u$};
  \node[box, fill=green!35]  (rcp) at ( 1.8, 3.7) {RC proj.\\$r$};

  \node[box, fill=orange!55] (utf) at (-1.8, 4.8) {$[u;\,r]$};
  \node[box, fill=green!55]  (rcf) at ( 1.8, 4.8) {$[r;\,u]$};

  \node[box, fill=orange!75] (uth) at (-1.8, 5.9) {UT head\\(10-class)};
  \node[box, fill=green!75]  (rch) at ( 1.8, 5.9) {RC head\\(4-class)};

  \draw[arr] (roberta) -- (tgt.south);

  \draw[arr] (tgt.north west) .. controls (-0.55, 1.9) .. (bos.south);
  \draw[arr] (tgt.north east) .. controls ( 0.55, 1.9) .. (eos.south);

  \draw[arr] (bos) -- (utp);
  \draw[arr] (eos) -- (rcp);
  \draw[arr] (utp) -- (utf);
  \draw[arr] (rcp) -- (rcf);
  \draw[arr] (utf) -- (uth);
  \draw[arr] (rcf) -- (rch);

  \draw[arr, dashed, gray] (rcp.north) .. controls (1.8,4.3) and (-0.4,4.3) .. (utf.east);
  \draw[arr, dashed, gray] (utp.north) .. controls (-1.8,4.3) and ( 0.4,4.3) .. (rcf.west);
\end{tikzpicture}
\caption{Dual-Probe Head (DPH) architecture. Each of the $2k{+}1$ context
utterances is individually wrapped as $\langle s\rangle d_i \langle/s\rangle$
before concatenation into the encoder input. For the target utterance $d_t$
(highlighted), the \texttt{<s>}~BOS hidden state $h_{\langle s\rangle}^{(t)}$
serves as the UT probe and the \texttt{</s>}~EOS hidden state
$h_{\langle/s\rangle}^{(t)}$ serves as the RC probe. Dashed arrows denote
cross-conditioning: the UT head receives $[u;\,r]$ and the RC head receives
$[r;\,u]$.}
\label{fig:arch}
\end{figure}
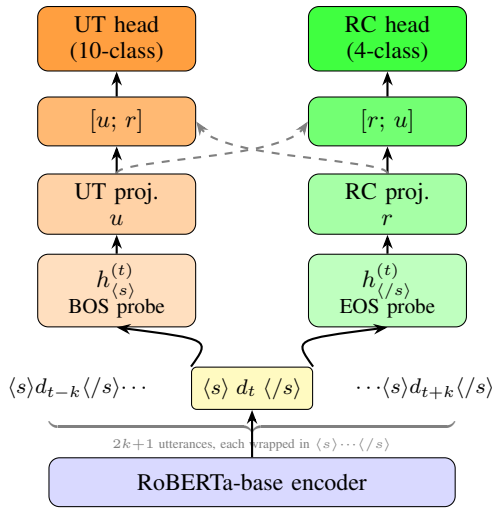

We develop a \emph{Dual-Probe Head} (DPH) model that performs joint
UT and RC classification. The architecture, illustrated in Fig.~\ref{fig:arch},
consists of three components.

\textbf{Encoder.} A shared RoBERTa-base backbone encodes the context
window. Each of the $W = 2k{+}1$ utterance slots is individually wrapped as
$\langle s\rangle d_i \langle/s\rangle$; the slots are concatenated to form a
single sequence. For the \emph{target}
utterance at the center slot, we read two positional \emph{probes} from the
encoder's last hidden layer: the \texttt{<s>}~(BOS) position yields
$h_{\langle s\rangle}^{(t)}\!\in\!\mathbb{R}^{768}$ (UT probe) and the
\texttt{</s>}~(EOS) position yields
$h_{\langle/s\rangle}^{(t)}\!\in\!\mathbb{R}^{768}$ (RC probe). Dedicating
distinct token positions to each task allows their attention patterns to
specialize independently during fine-tuning.

\textbf{Task projections.} Two linear layers project the respective probes:
$u = \text{ReLU}(W_u\,h_{\langle s\rangle}^{(t)})$ and
$r = \text{ReLU}(W_r\,h_{\langle/s\rangle}^{(t)})$.

\textbf{Cross-conditioned heads.} The UT head receives the concatenated feature
$[u; r]$ and the RC head receives $[r; u]$. This symmetric
conditioning allows each task to exploit the other's intermediate representation,
capturing the correlation between utterance function and reasoning quality without
a sequential prediction bottleneck. 

\textbf{Loss.} We use focal loss~\cite{Lin2017-en} ($\gamma=2$) with
inverse-frequency class weights for both tasks. The joint training loss is
$\mathcal{L} = \alpha \mathcal{L}_{UT} + (1-\alpha)\mathcal{L}_{RC}$ with
$\alpha = 0.5$. $\alpha$ is determined via grid search on the validation set,
prioritizing the primary metric. The bottom two-thirds of RoBERTa layers are
frozen during training to reduce the risk of catastrophic forgetting on this
small dataset.

\textbf{Training details.} Optimizer: AdamW; learning rate $\in \{$1e-5, 2e-5,
3e-5$\}$; batch size 16; early stopping on validation macro-F1 with patience 3.

\subsection{Baselines}

We compare against three systems:

\begin{enumerate}
  \item \textbf{TF-IDF + Logistic Regression}: Bag-of-words feature vector with
    $\ell_2$-regularized logistic regression (class-balanced weights).
  \item \textbf{LLM Zero-shot GPT-5.4}: Prompt containing the full coding-scheme
    definitions and 3 in-context examples per class; no fine-tuning.
  \item \textbf{RoBERTa-base without augmentation}: DPH trained on original
    labeled data only.
\end{enumerate}

\subsection{Evaluation Protocol}

\textbf{Primary metric:} Macro-F1, which averages F1 across all classes equally
and prevents the dominant SR-D class from masking poor minority-class performance.

\textbf{Secondary metrics:} (Macro-/Weighted-) precision, recall, and F1;
overall accuracy; confusion matrix.

\textbf{Validation strategy:} The session-level exhaustive-search 6/2/1 split
described in Section~\ref{sec:data} is used throughout. The fixed validation
set (2 sessions) is used for hyperparameter selection and early stopping; the
held-out test set (1 session, $N=305$ utterances) is used for final unbiased
reporting. Statistical significance of the best model vs.\ strongest baseline
is assessed with McNemar's test ($\alpha = 0.05$).

\subsection{Discourse Pattern Analysis}

\textbf{Co-occurrence analysis.} We construct UT$\times$RC contingency tables
and compute the row-normalized conditional probability $P(\text{RC} | \text{UT})$
to identify which utterance types are associated with which reasoning levels.

\textbf{Cognitive Complexity Index (CCI).} For each session, we compute the mean
CCI across all utterances (and separately for student utterances only). Sessions
are compared using the CCI as a scalar measure of discourse cognitive depth.

\textbf{Temporal analysis.} Each session is divided into $n$ equal time bins.
CCI is computed per bin to reveal within-session cognitive trajectories.

\textbf{Lag-sequential analysis.} For each teacher utterance at turn $t$, we
record the RC label of the immediately following student utterance at turn $t+1$.
This yields a transition table $P(\text{RC}_{\text{student}} | \text{UT}_{\text{teacher}})$
that quantifies which teacher moves are most effective at eliciting higher-order
student reasoning.

\textbf{Instructional move sequence (IRF chain) analysis.} We extend the lag-1
analysis to full 3-turn chains to see if we can find an IRF-like pattern of teacher
move $\rightarrow$ student reasoning $\rightarrow$ teacher feedback that is
associated with higher student CCI. All consecutive
Teacher-Student-Teacher (T-S-T) triples within each session are extracted; for each
triple the UT 3-gram $(ut_{t-1},\,ut_t,\,ut_{t+1})$ and the student CCI at position
$t$ are recorded. The 26 patterns with $n \geq 5$ occurrences are ranked by mean
student CCI.
A teacher-framing heatmap then aggregates mean student CCI over all
(initiation UT$\times$feedback UT) pairs, collapsing the student UT dimension to
isolate the teacher-controlled portion of the IRF frame. Finally, we compute
$P(\text{Rq} | \text{UT}_{\text{teacher}})$ to identify which teacher moves
create discursive space for student-initiated inquiry.

\section{Results}

\subsection{Classification Results}
\label{sec:classification}

\begin{table}[t]
\centering
\caption{Utterance Type (UT) classification results across models.
  Macro-F1 is the primary metric. ``+Aug'' indicates training with
  LLM-generated synthetic data.}
\label{tab:results-ut}
\begin{tabular}{lcccc}
\toprule
\textbf{Model} & \textbf{Acc} & \textbf{Mac-P} & \textbf{Mac-R} & \textbf{Mac-F1} \\
\midrule
TF-IDF + Logistic Regression    & 0.331 & 0.329 & 0.359 & 0.292 \phantom{$\dagger$} \\
Zero-shot GPT-5.4               & 0.620 & 0.626 & 0.783 & 0.582 \phantom{$\dagger$} \\
RoBERTa-base (no aug)           & 0.579 & 0.536 & 0.631 & 0.543 \phantom{$\dagger$} \\
RoBERTa-base (+Aug)             & 0.683 & 0.628 & 0.681 & \textbf{0.635} $\dagger$ \\
\bottomrule
\multicolumn{5}{l}{\footnotesize $\dagger$ McNemar's test (vs.\ TF-IDF+LR): $p < 0.001$} \\
\end{tabular}
\end{table}

\begin{table}[t]
\centering
\caption{Reasoning Component (RC) classification results across models (4-class
scheme: ER, SR-D, SR-I, O). Macro-F1 is the primary metric. ``+Aug'' indicates
training with LLM-generated synthetic examples.}
\label{tab:results-rc}
\begin{tabular}{lcccc}
\toprule
\textbf{Model} & \textbf{Acc} & \textbf{Mac-P} & \textbf{Mac-R} & \textbf{Mac-F1} \\
\midrule
TF-IDF + Logistic Regression    & 0.685 & 0.380 & 0.388 & 0.381 \phantom{$\dagger$} \\
Zero-shot GPT-5.4               & 0.577 & 0.450 & 0.569 & 0.412 \phantom{$\dagger$} \\
RoBERTa-base (no aug)           & 0.477 & 0.419 & 0.422 & 0.345 \phantom{$\dagger$} \\
RoBERTa-base (+Aug)             & 0.649 & 0.395 & 0.446 & 0.396 \phantom{$\dagger$} \\
\bottomrule
\end{tabular}
\end{table}


Tables~\ref{tab:results-ut} and~\ref{tab:results-rc} report classification
performance for all models.

TF-IDF + Logistic Regression presents a split picture across tasks.  On UT it
achieves macro-F1$=$0.292 — the lowest among all models — consistent with a
bag-of-words approach failing on context-dependent functional distinctions
(e.g., Fq vs.\ Q, P vs.\ E).  On RC, it achieves a competitive macro-F1
of 0.381 (Acc$=$0.685), ranking third among the four models, driven by strong
SR-D (F1$=$0.766) and SR-I (F1$=$0.583) performance.  This suggests that the
4-class RC taxonomy has meaningful lexical separability: utterances containing
scientific terminology reliably predict SR-D, while pattern/inference language
predicts SR-I, enabling a bag-of-words model to perform competitively despite
lacking contextual representations.

Zero-shot GPT-5.4 achieves macro-F1$=$0.582 on UT and macro-F1$=$0.412 on RC.
On UT, the model performs well on high-frequency classes (Ry: F1$=$0.800;
E: F1$=$0.581; Q: F1$=$0.549) but struggles with rare or functionally
overlapping classes (P: F1$=$0.229; Rq: F1$=$0.375).  On RC, GPT-5.4
achieves solid performance on O (F1$=$0.671) and improved minority-class
coverage (ER: F1$=$0.444), though SR-I remains difficult (F1$=$0.242).

Fine-tuning RoBERTa-base on the labeled corpus alone substantially improves
UT performance (macro-F1$=$0.543) but does not match TF-IDF on RC
(macro-F1$=$0.345).  Adding LLM-based data augmentation at scale $\times$3.0
yields the best UT result overall: macro-F1$=$0.635 ($+$34.3 pp over
TF-IDF), with gains distributed across minority classes including Fq, Ry, and
ER.  On RC, augmentation improves RoBERTa to 0.396, narrowly exceeding TF-IDF
(0.381) by 1.5 pp; however, GPT-5.4 retains the highest RC macro-F1 (0.412)
across all systems.


\subsubsection{Confusion Matrix}
\begin{figure}[t]
  \centering
  \includegraphics[width=.8\linewidth]{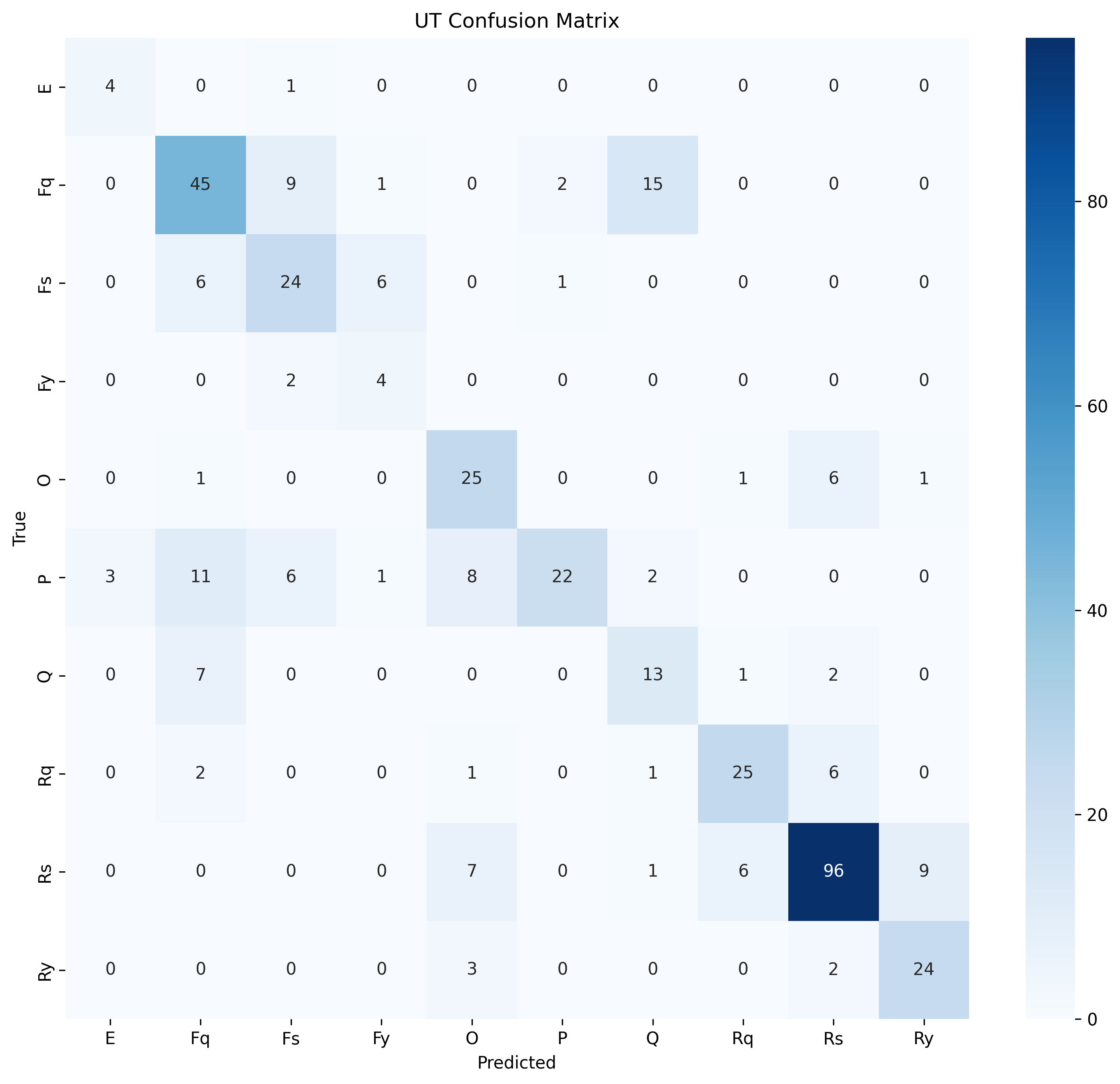}
  \caption{Confusion matrix for Utterance Type (UT) classification
    (RoBERTa-base +Aug, 10-class).  Rows are true labels; columns are
    predicted labels.}
  \label{fig:ut-cm}
\end{figure}

Figure~\ref{fig:ut-cm} shows the confusion matrix for the best-performing UT
model (RoBERTa-base +Aug). The most prominent source of error is the Prompt (P)
class, reflecting that the surface-level similarity between teacher prompt
moves and other utterance types, often containing question-like or facilitative
language. The second most common confusion runs between Fq and Q, showing the
functional overlap between teacher follow-up questions and student-generated
questions in dialogic exchanges. 

For the RC task (4-class: ER, O, SR-D, SR-I), the dominant error pattern
involves the two highest-order reasoning categories: 26 of 31 SR-I
(Inferential) instances (83.9\%) are predicted as SR-D (Descriptive), yielding
near-zero SR-I recall. Given that the SR-I class is the first minority class,
the model might learn a decision boundary skewed toward SR-D to make more
conservative prediction. Also, we observe a class confusion between O (Other)
and SR-D (Descriptive), with 32 O instances (19.0\%) classified as SR-D, and 40
SR-D instances (19.2\%) classified as O. This suggests that descriptive
reasoning statements often share surface features with off-topic or procedural
talk, leading to misclassification. 

\begin{table}[t]
\centering
\caption{Effect of LLM-based data augmentation on UT and RC classification
  (RoBERTa-base). Macro-F1 is the primary metric; best result per task
  is \textbf{bold}.}
\label{tab:aug-ablation}
\begin{tabular}{lcccc}
\toprule
 & \multicolumn{2}{c}{\textbf{UT}} & \multicolumn{2}{c}{\textbf{RC}} \\
\cmidrule(lr){2-3}\cmidrule(lr){4-5}
\textbf{Training Data} & \textbf{Acc} & \textbf{Mac-F1} & \textbf{Acc} & \textbf{Mac-F1} \\
\midrule
Original only (no aug)     & 0.579 & 0.543 & 0.477 & 0.345 \\
+Minority boost            & 0.596 & 0.522 & 0.487 & 0.308 \\
+Synthetic ($\times$0.5)   & 0.659 & 0.610 & 0.523 & 0.362 \\
+Synthetic ($\times$1.0)   & 0.668 & 0.621 & 0.535 & 0.362 \\
+Synthetic ($\times$1.5)   & 0.673 & 0.615 & 0.593 & 0.384 \\
+Synthetic ($\times$2.0)   & 0.666 & 0.604 & 0.555 & 0.374 \\
+Synthetic ($\times$2.5)   & 0.683 & 0.633 & 0.564 & 0.361 \\
+Synthetic ($\times$3.0)   & \textbf{0.683} & \textbf{0.635} & \textbf{0.649} & \textbf{0.396} \\
+Synthetic ($\times$3.5)   & 0.668 & 0.587 & 0.617 & 0.388 \\
\bottomrule
\end{tabular}
\end{table}

\subsection{Discourse Pattern Analysis}
\label{sec:discourse}

\subsubsection{UT$\times$RC Co-occurrence}

\begin{figure*}[!t]
\centering
\includegraphics[width=.9\linewidth]{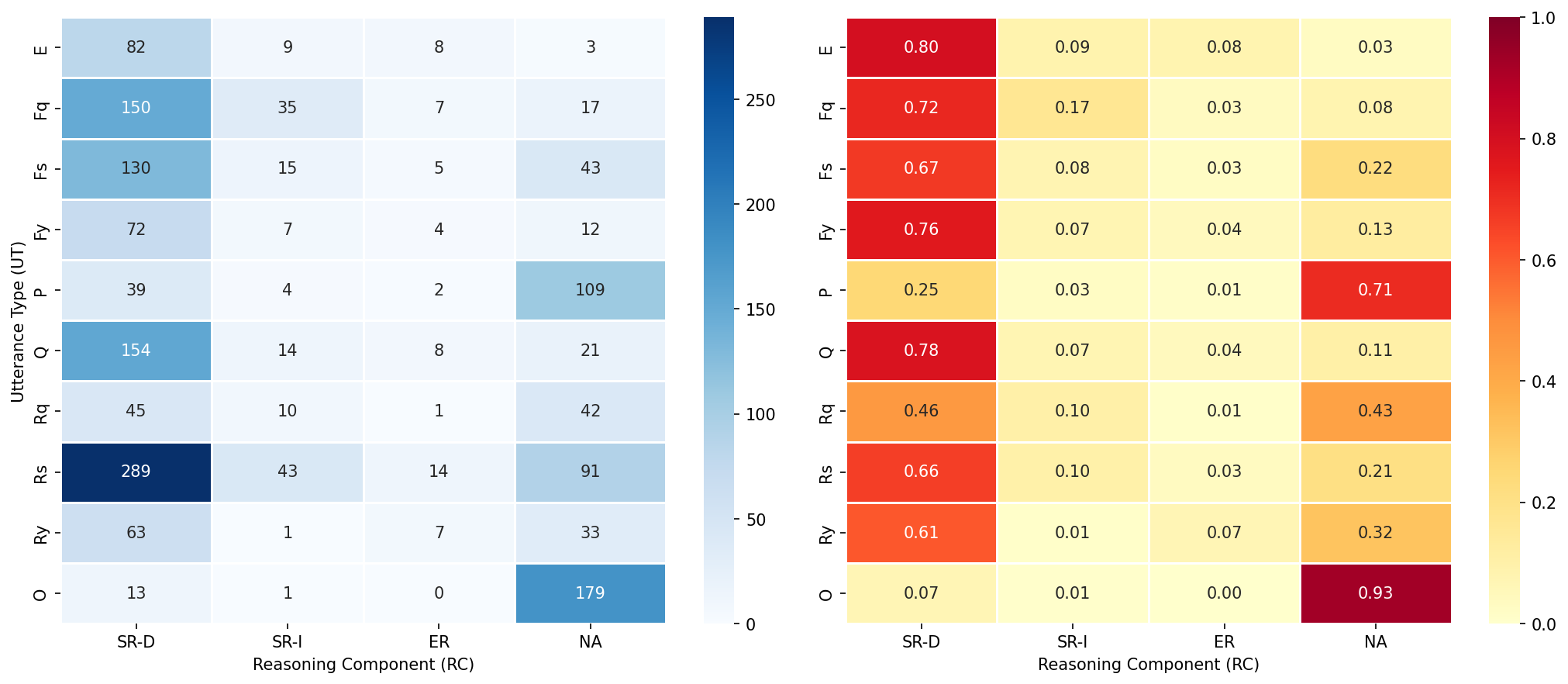}
\caption{Utterance Type (UT)$\times$Reasoning Component (RC) Co-occurrence Heatmap. Left: raw counts; Right:
$P(\text{RC} | \text{UT})$ (row-normalized).}
\label{fig:cooccurrence}
\end{figure*}

Fig.~\ref{fig:cooccurrence} shows the UT$\times$RC co-occurrence heatmap.
Row-normalized values reveal systematic patterns: teacher Questions (Q) are
predominantly paired with SR-D responses ($P(\text{SR-D}|\text{Q}) \approx 0.68$),
consistent with teachers using recall questions to elicit factual scientific content.
Feedback-with-Question (Fq) is the utterance type most associated with SR-I
student reasoning ($P(\text{SR-I}|\text{Fq}) \approx 0.15$), nearly twice the
rate following plain Questions. Prompts (P) have the highest proportion of NA
student responses, suggesting that open-ended prompts are often met with
procedural or off-topic replies. Student Responses-as-Statements (Rs) are
almost exclusively paired with SR-D, confirming their role as the primary vehicle
for scientific knowledge recitation.

%
%
%
\subsubsection{Temporal Analysis}
\label{sec:temporal}

\begin{figure}[t]
\centering
\includegraphics[width=\columnwidth]{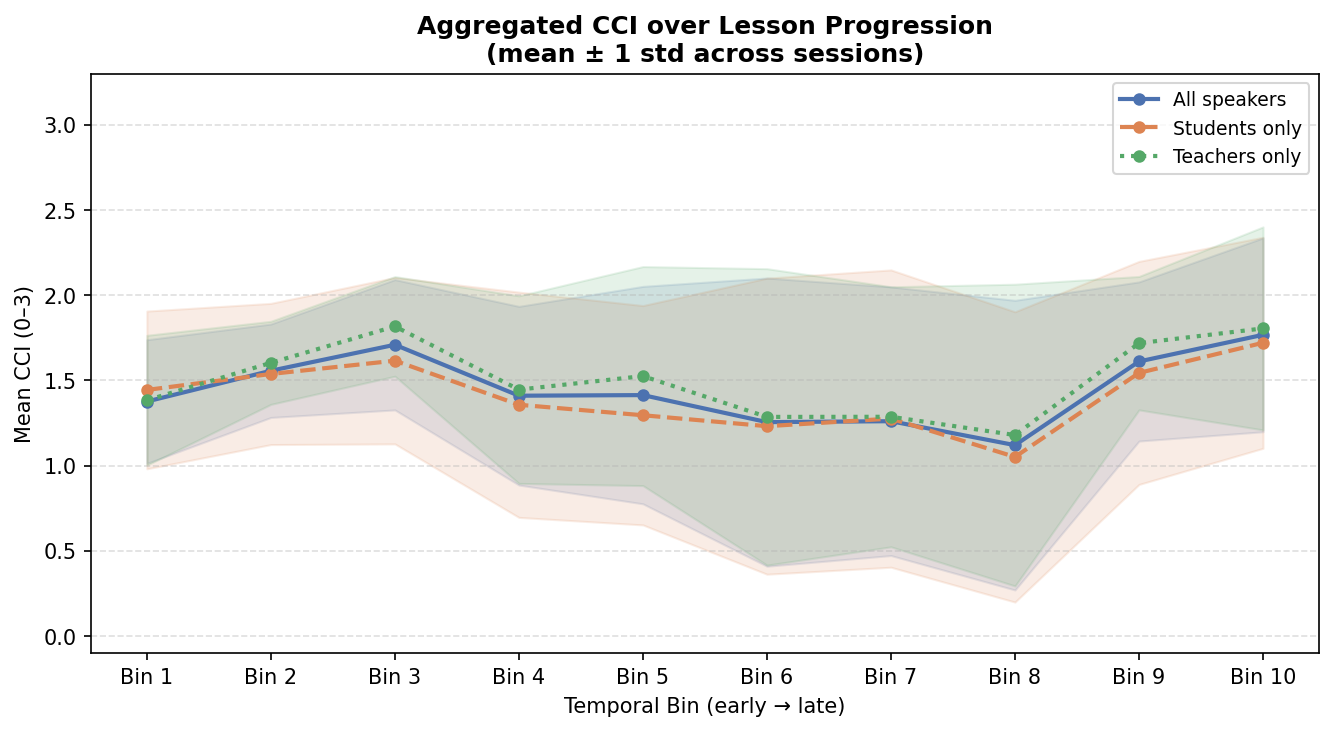}
\caption{Aggregated CCI Time Series across Lesson Progression (binned into 10 equal
segments). Shaded region indicates $\pm$1 standard deviation.}
\label{fig:cci_time}
\end{figure}

Fig.~\ref{fig:cci_time} shows the within-session CCI trajectory aggregated
across all sessions. A modest but consistent rise in mean CCI is observed from
the beginning of sessions, followed by a slight decline toward the end, and a
sharp rebound occurs towards the end of session where mean CCI returns to or
exceeds the beginning-session peak. This pattern is consistent with the
pedagogical structure of Initiation (lower-order recall) $\to$ Development
(higher-order analysis) $\to$ Procedural (low-level observations and
management) $\to$ Synthesis (reflective closure) where teachers reassert
cognitive press at the end of the lesson to consolidate conceptual
understanding. The mid-session dip may reflect a procedural phase where
students are engaged in hands-on activities or data collection, which may
involve more descriptive talk and less inferential reasoning.

\subsubsection{Lag-Sequential Analysis}

Table~\ref{tab:lag_stats} reports the distribution of student RC codes following
each teacher UT type. Two patterns are particularly striking. First, Fq (Feedback
with Question) is the teacher move most consistently followed by SR-I student
reasoning (14.6\%).
This is more than twice the SR-I rate following plain Questions (7.3\%). Second,
teacher Explanations (E) are rarely followed by SR-I (0 occurrences), suggesting
that extended teacher monologue suppresses rather than elicits student inferential
engagement.

\begin{table}[t]
\centering
\caption{Lag-sequential analysis: distribution of student RC at turn $t{+}1$
following each teacher UT at turn $t$. Bold: highest SR-I count per row.}
\label{tab:lag_stats}
\begin{tabular}{lrrrr|r}
\toprule
\textbf{Teacher UT} & \textbf{SR-D} & \textbf{SR-I} & \textbf{ER} & \textbf{O} & \textbf{Total} \\
\midrule
E  (Explanation)    & 8   & 0            & 2   & 11  & 21  \\
Fq (Feedback-Q)     & 126 & \textbf{29}  & 8   & 36  & 199 \\
Fs (Feedback-S)     & 54  & 0            & 2   & 49  & 105 \\
Fy (Feedback-Y)     & 14  & 2            & 1   & 15  & 32  \\
P  (Prompt)         & 25  & 2            & 2   & 66  & 95  \\
Q  (Question)       & 113 & \textbf{12}  & 7   & 33  & 165 \\
\bottomrule
\end{tabular}
\end{table}

\subsubsection{Instructional Move Sequence Analysis}

\begin{figure}[t]
\centering
\includegraphics[width=\columnwidth]{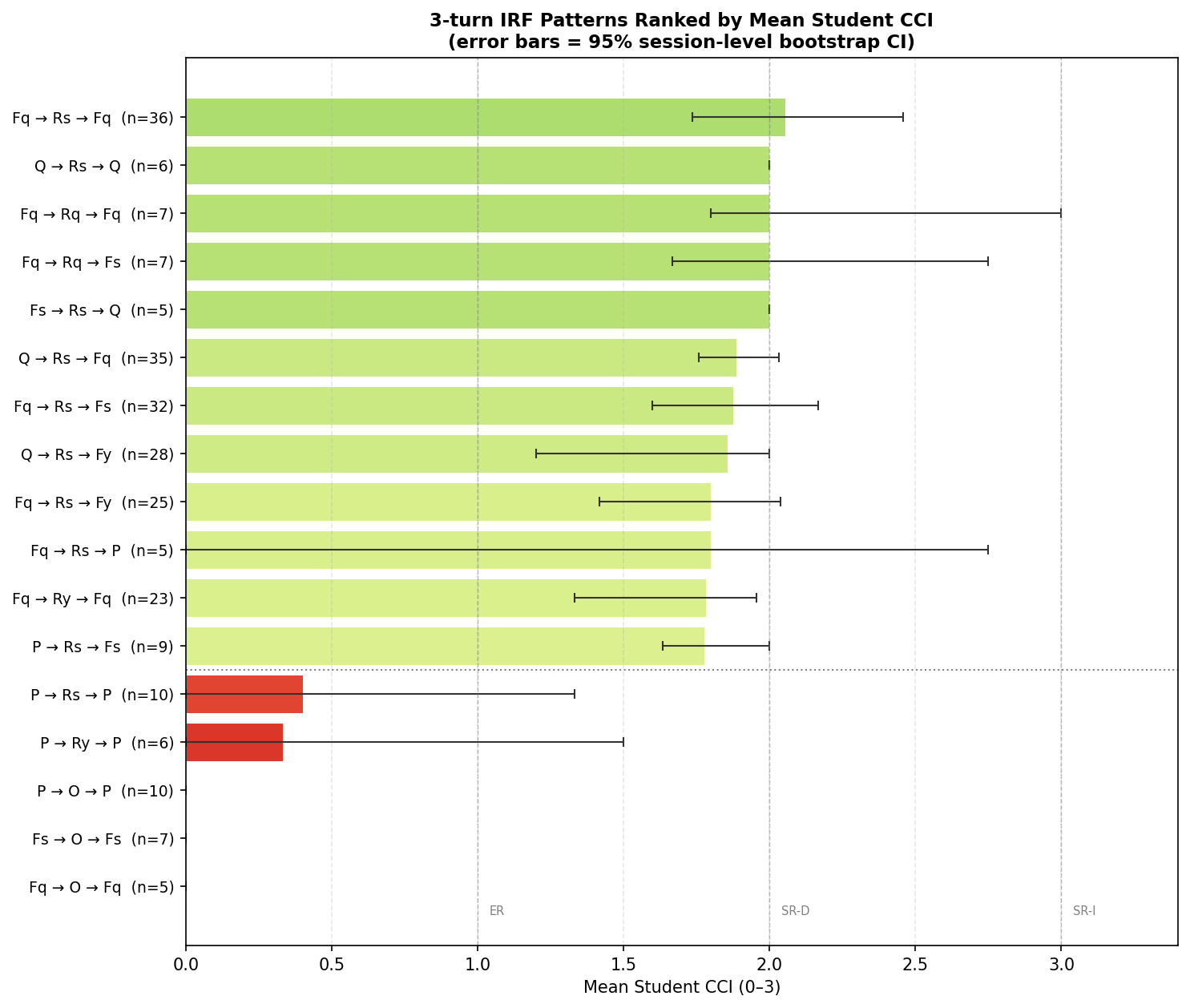}
\caption{3-Turn IRF Patterns Ranked by Mean Student CCI (min. $n$=5, $N$=26
patterns). Error bars are session-level bootstrap 95\% CIs. Bar color encodes
CCI magnitude (red$\to$green scale). Top-ranked: \textbf{Fq$\to$Rs$\to$Fq}
($n$=36, mean CCI=2.06).}
\label{fig:3gram_bar}
\end{figure}

\begin{figure}[t]
\centering
\includegraphics[width=\columnwidth]{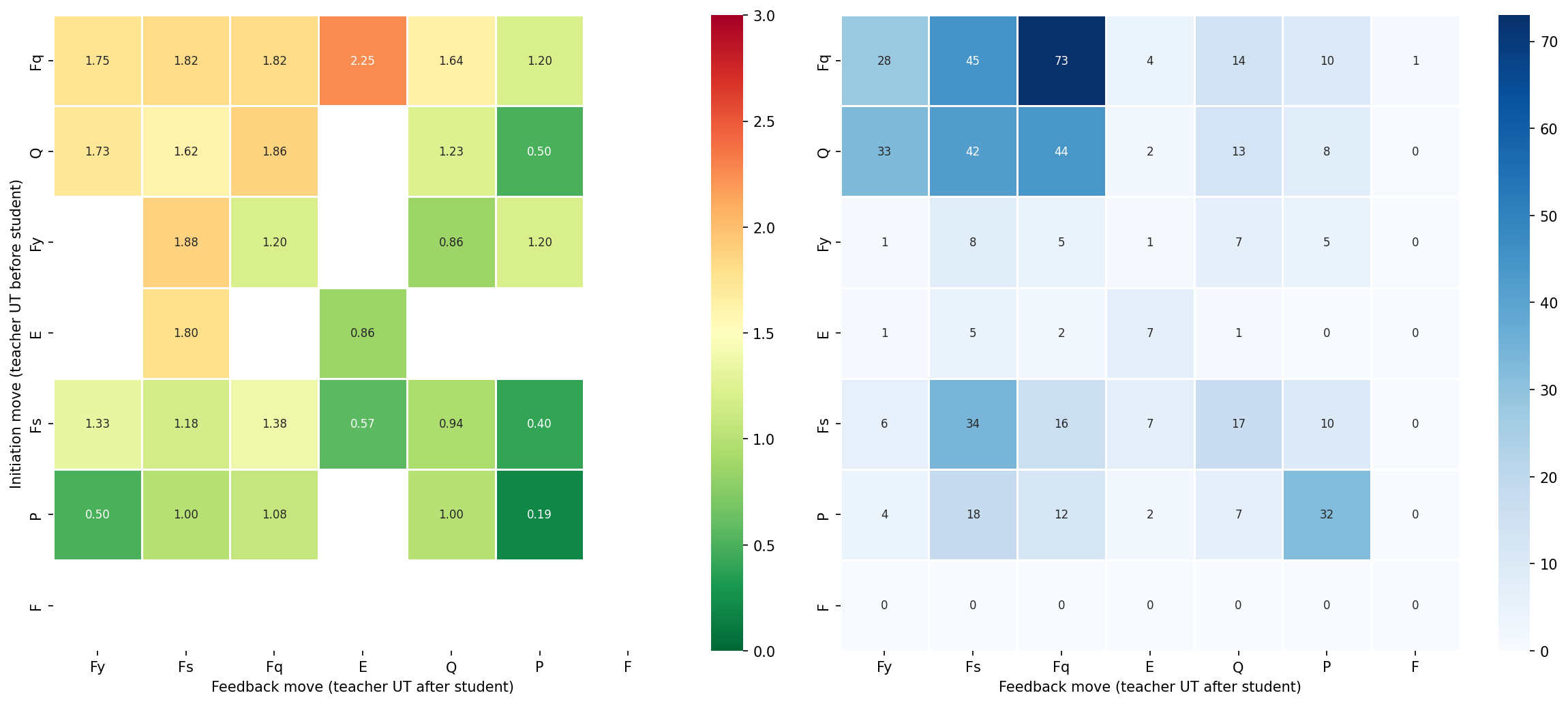}
\caption{Teacher-framing heatmap. Left: mean student CCI per
(initiation UT$\times$feedback UT) pair (grey $=$ fewer than 3 observations).
Right: observation count per cell.}
\label{fig:framing_heatmap}
\end{figure}

\begin{figure}[t]
\centering
\includegraphics[width=\columnwidth]{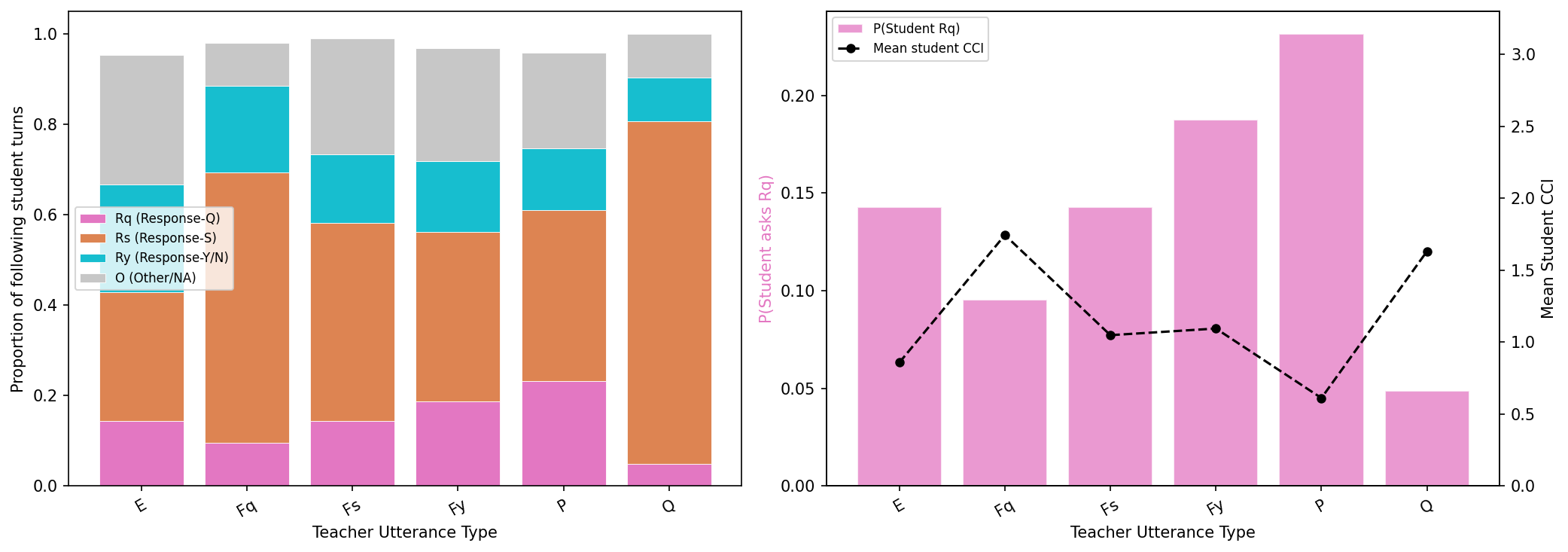}
\caption{Rq trigger analysis. Left: $P(\text{student UT} | \text{teacher UT})$
for all lag-1 T$\to$S transitions ($N=623$ pairs). Right: $P(\text{student Rq})$
(bars) and mean student CCI (line) per preceding teacher UT.}
\label{fig:rq_triggers}
\end{figure}

Fig.~\ref{fig:3gram_bar} shows all 3-turn Teacher-Student-Teacher (T-S-T)
patterns with at least 5 observations, ranked by mean student CCI. The
top-ranked pattern is \textbf{Fq$\to$Rs$\to$Fq} ($n$=36, mean CCI=2.06, 95\% CI
$[1.74,\,2.46]$), in which a teacher feedback-question elicits a student
declarative response that is itself followed by a further feedback-question.
Two other patterns reach mean CCI=2.00: \textbf{Q$\to$Rs$\to$Q} ($n$=6) and
\textbf{Fq$\to$Rq$\to$Fq} ($n$=7), the latter notably involving a student
question in the middle position. Together these top-ranked patterns share the
structural property that the teacher re-initiates with a question after the
student's turn, regardless of student response type. At the bottom,
\textbf{P$\to$O$\to$P} ($n$=10, mean CCI=0.00) and two analogous prompt-flanked
patterns with inaudible/off-topic student turns show no reasoning engagement
whatsoever.

The teacher-framing heatmap (Fig.~\ref{fig:framing_heatmap}) isolates the
teacher-controlled component of the 3-turn instructional move by collapsing the
student UT dimension.  The single highest-CCI framing is \textbf{Fq$\times$Q}
(mean CCI=2.23, $n$=14): when a teacher initiates with a follow-up question
\emph{and} the student's high inferential response is met with a further
question. Although this framing is relatively rare, the sharp CCI peak suggests
that sustained question-on-question pressure is a particularly potent scaffold
for inferential reasoning.  The most \emph{frequent} high-CCI framings are
\textbf{Fq$\times$Fq} (73 windows, mean CCI$=$1.82) and \textbf{Fq$\times$Fs}
(45 windows, mean CCI$=$1.82), confirming that Fq-initiated frames reliably
elicit above-average student reasoning regardless of whether the feedback move
is another question or a brief follow-up statement.  \textbf{Q$\times$Fq} (44
windows, mean CCI$=$1.96) similarly ranks among the high-CCI framings,
reinforcing the pattern that any question to students promotes higher
reasoning, and student's inferential responses also encourage teachers to
follow up with questions. The most common low-CCI framing remains
\textbf{P$\times$P} (32 windows, mean CCI$=$0.18): consecutive prompt moves
without escalation to a question consistently fail to draw out inferential
student reasoning.

Fig.~\ref{fig:rq_triggers} reports $P(\text{Rq} | \text{UT}_{\text{teacher}})$
alongside mean student CCI per preceding teacher UT. Teacher \textbf{Prompts (P)}
produce the highest probability of a student question-asking response
($P(\text{Rq})=0.23$), followed by confirmatory feedback Fy ($0.19$). By
contrast, teacher \textbf{Questions (Q)} strongly suppress Rq ($P(\text{Rq})=0.05$)
while maximizing direct student statements ($P(\text{Rs})=0.76$). This reveals a
trade-off: Fq and Q moves maximize student CCI but minimize student-initiated
inquiry, while P and Fy moves create discursive space for students to ask
questions at the cost of lower mean CCI.

\subsection{Human-Labeled vs.\ Pseudo-Labeled Analysis}
\label{sec:pseudo}

To assess the extent to which discourse-level findings generalize beyond the
annotated corpus, we replicated the key analyses on a larger dataset labeled by
the best-performing RoBERTa-base (+Aug) model. We sampled 10 sessions from the
unlabeled data, applied the trained model to generate UT and RC predictions for
all utterances. Full comparison tables are provided in the Appendix; here we
highlight the most theoretically consequential divergence between the two label
sources.

\paragraph{CCI Temporal Trajectory.}

\begin{figure}[t]
  \centering
  \includegraphics[width=\columnwidth]{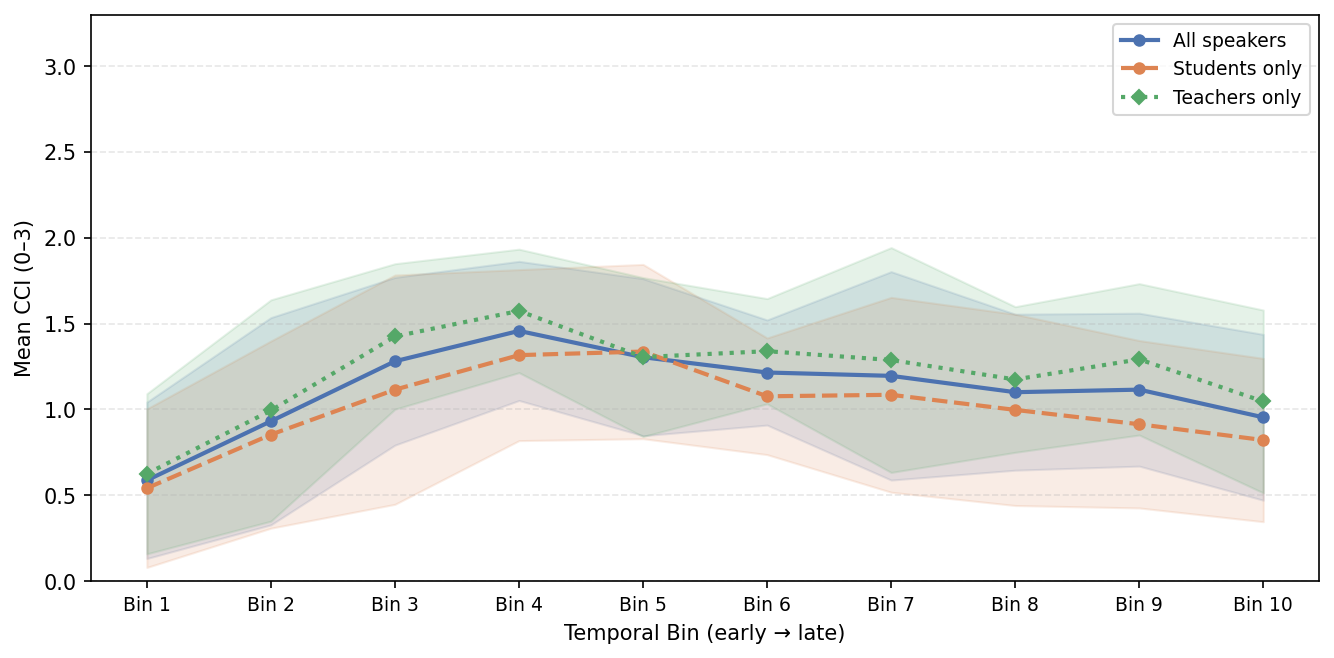}
  \caption{Aggregated CCI over lesson progression computed on
    pseudo-labeled data (model predictions). Compare with
    Fig.~\ref{fig:cci_time} (human-labeled): the sharp terminal rebound
    visible in the human-labeled trajectory is absent here.}
  \label{fig:cci_time_pseudo}
\end{figure}

The most surprising discrepancy emerges in the within-session CCI trajectory
(Fig.~\ref{fig:cci_time}).  In the human-labeled data the trajectory exhibits
a sharp rebound in the final two temporal bins (Bins 9--10), as described in
Section~\ref{sec:temporal}.  This terminal uptick is \emph{absent} in the
pseudo-labeled data: as shown in Fig.~\ref{fig:cci_time_pseudo}, the
model-predicted CCI trajectory rises modestly through Bin 4 and then declines
monotonically through Bin 10, with no late-session recovery.

We attribute this divergence to a form of \textbf{positional annotation bias}
in the human-labeled data. Human annotators read each utterance with full
awareness of its position in the lesson: an utterance at the end of a session
is contextualized as a closing synthesis or reflective summary, and annotators
likely assign higher-order RC codes (SR-I, SR-D) accordingly.  The trained
model, by contrast, operates strictly at the utterance level and has no
representation of lesson-positional context. Detailed per-session comparisons
and additional discourse metrics under both label sources are reported in
Appendix~\ref{app:pseudo_comparison}.

\section{Discussion}

\subsection{On the Revised Coding Scheme}

The collapse from six to four RC classes is a pragmatic but theoretically principled
decision. The key insight from the CDAT framework is that SK and OD are both
\emph{descriptive} scientific acts --- one recalls established knowledge, the other
reports observed phenomena --- while PD and MT are both \emph{inferential} acts that
draw conclusions beyond the data. The merged scheme (SR-D, SR-I) preserves this
epistemically meaningful distinction while yielding learnable class sizes and
reducing annotator boundary ambiguity.

\subsection{On Zero-Shot vs.\ Fine-Tuned Performance}

The results reveal a sharp task-dependent divergence.  On UT, fine-tuning
with augmentation yields the best macro-F1 ($+$34.3 pp over TF-IDF), and
GPT-5.4 also achieves a large gain over TF-IDF ($+$29.0 pp): the 10-class
functional taxonomy requires discourse context that bag-of-words
representations cannot adequately exploit, and both LLMs and fine-tuned
transformers capture this contextual information effectively.

On RC, GPT-5.4 achieves the highest macro-F1 (0.412), followed closely by
RoBERTa+Aug (0.396) and TF-IDF (0.381), with RoBERTa without augmentation
performing worst (0.345).  The narrow gap among the top three systems
reflects the meaningful lexical separability of the 4-class RC taxonomy:
SR-D and SR-I are partially predicted by scientific vocabulary and inference
language respectively, enabling competitive performance even from a
bag-of-words model.  Fine-tuned RoBERTa without augmentation underperforms on
RC, likely because the small corpus ($\approx$985 training examples) is
insufficient to overcome RoBERTa's tendency to overfit; LLM-based
augmentation closes most of this gap, bringing RoBERTa+Aug to within 1.5 pp
of TF-IDF and 1.6 pp of GPT-5.4.  This finding motivates corpus expansion as
a priority for future work: with substantially more labeled data, fine-tuned
transformers are likely to pull ahead more decisively on RC as well.

\subsection{On Prompt Design: Whole-Snippet Generation.}

An important design decision concerns the unit of generation.  Our initial
prompt asked the model to vary \emph{only the target utterance} while holding
the surrounding context window fixed, with the ground-truth UT and RC labels
provided as conditioning.  This yielded no measurable performance gain. We
conjecture that the model learned to exploit the fixed context as a shortcut
cue for label prediction, rather than learning to generalize from the varied
target utterance.

We therefore redesigned the prompt to regenerate \emph{the entire discourse
snippet} --- all turns in the context window plus the target --- as a coherent
unit.  Each variation thus differs both in the target utterance \emph{and} in
the surrounding conversational context, preventing any single contextual
pattern from becoming a reliable label cue.  Empirically, this change produced
consistent improvements in macro-F1 across augmentation scales
(Table~\ref{tab:aug-ablation}), validating the design choice.

\subsection{On Discourse Patterns and Instructional Practice}

The lag-sequential and IRF chain results (Section~\ref{sec:discourse}) converge
on a single actionable principle: the teacher move that most reliably elevates
student reasoning is the Feedback-with-Question (Fq), which acknowledges a
student contribution and immediately presses for deeper reasoning --- precisely
the ``press for reasoning'' move in Accountable Talk \cite{Michaels2008-ys}.
The \textbf{P$\times$P} framing identified in the teacher-framing heatmap is a
high-prevalence, high-impact counterexample and represents a concrete target for
professional development intervention.

The Rq trigger analysis surfaces a substantive pedagogical trade-off: the move
types that maximize student CCI (Fq, Q) are also those that most suppress
student-initiated inquiry, while the move that most invites student questions (P)
yields the lowest reasoning quality when used repetitively. Effective teachers
may need to sequence both types strategically --- Fq/Q for reasoning depth, P/Fy
to open inquiry space.

The temporal trajectory comparison between human-labeled and pseudo-labeled data
(Section~\ref{sec:pseudo}) further reveals that the end-of-lesson CCI rebound
visible in human annotations reflects positional bias, a finding that suggest
model improvements that incorporate lesson-level context may be necessary to
capture this pattern. This also raises a cautionary flag for interpreting the
temporal CCI pattern as a genuine pedagogical phenomenon.

\subsection{Limitations}

Several limitations bound the generalizability of our findings. First, the
corpus is small ($\approx$1,782 labeled utterances from a limited number of
middle-school sessions) and may not represent science classroom discourse more
broadly, which are also heavily dominated by mathematics lessons. Second,
inter-annotator agreement on the revised 4-class scheme has not yet been
formally computed; this is a necessary validation step before the scheme can be
recommended for general use. 

\section{Conclusion}

We presented an automated discourse analysis system for science classroom utterances
that jointly classifies utterance type (10 classes) and reasoning component
(revised 4-class scheme). Our findings demonstrate that:

\begin{itemize}
  \item LLM-based augmentation of minority RC classes is a practical strategy for
    improving automated classification of rare but educationally significant reasoning
    types, with gains concentrated on SR-I and ER F1 scores.
  \item Fine-tuned RoBERTa with augmentation achieves the best UT performance
    (macro-F1$=$0.635, $+$34.3 pp over TF-IDF), confirming that 10-class
    functional discourse labeling requires contextual representation.  By
    contrast, RC performance is competitive across all systems: GPT-5.4
    achieves the highest RC macro-F1 (0.412), followed by RoBERTa+Aug (0.396)
    and TF-IDF (0.381), indicating that the 4-class RC taxonomy has meaningful
    lexical separability and that transformer fine-tuning on the current corpus
    size does not provide a decisive advantage for this task.
  \item Teacher Feedback-with-Question (Fq) moves are the strongest antecedents of
    student inferential reasoning (SR-I) in the lag-sequential analysis, providing
    quantitative support for dialogic teaching practices that combine feedback with
    epistemic probing.
  \item Cognitive complexity follows a within-session arc that peaks in the lesson
    middle, with implications for how teachers structure the closing phases of
    science discussions.
\end{itemize}

Future work will expand the annotated corpus and report inter-annotator
reliability. Also, we plan to investigate the reliable methods to measure the
uncertainty of the model predictions and active learning strategies to
efficiently expand the labeled dataset with human-in-the-loop. 

\bibliographystyle{IEEEtran}
\bibliography{references}
\appendices

\section{Synthetic Data Generation Prompt}
\label{app:prompt}

The following prompt template was used to generate synthetic dialogue variations
via GPT-4.1. Placeholders in curly braces are filled at runtime with the
corresponding values from each training example.

\begin{quote}
\ttfamily\small\obeylines\obeyspaces
\textbf{<role>}\\
You are a classroom discourse analyst specializing in science education dialogue patterns.\\
\textbf{</role>}

\textbf{<task>}\\
Generate \{variations\} distinct variations of a science classroom dialogue snippet.\\
Each variation must re-write the ENTIRE snippet (all \{window\} turns), not just the target utterance.\\
The target utterance is turn \{center\_1indexed\} (marked with ***).\\
All other turns must flow naturally and coherently with the new target utterance.\\
\textbf{</task>}

\textbf{<label\_definitions>}\\
\{label\_definitions\}\\
\textbf{</label\_definitions>}

\textbf{<original\_snippet>}\\
\{original\_snippet\}\\
\textbf{</original\_snippet>}

\textbf{<requirements>}\\
1. Generate exactly \{variations\} complete snippet variations\\
2. In each variation, the target turn (turn \{center\_1indexed\}) must preserve its discourse labels (UT: \{label\_ut\}, RC: \{label\_rc\}) and speaker (\{speaker\})\\
3. All surrounding turns must be rewritten so the dialogue flows naturally with the new target; do NOT copy surrounding turns verbatim from the original\\
4. Use different vocabulary, phrasing, and sentence structures across variations\\
5. Maintain scientific accuracy and classroom-appropriate language\\
6. Respond with ONLY valid JSON --- no prose, no markdown fences\\
\textbf{</requirements>}

\textbf{<output\_format>}\\
{[}\\
\hspace*{1em}\{\\
\hspace*{2em}``before'': [``turn 1 text'', ``turn 2 text'', \ldots],\\
\hspace*{2em}``target'': ``target turn text'',\\
\hspace*{2em}``after'':  [``turn N+1 text'', ``turn N+2 text'', \ldots]\\
\hspace*{1em}\},\\
\hspace*{1em}\ldots\\
{]}\\
\textbf{</output\_format>}
\end{quote}

The \texttt{label\_definitions} block enumerates all UT and RC labels with their
names and definitions. The \texttt{original\_snippet} block lists each turn in
the context window in the form \texttt{[i] Speaker: ``text''}, with the target
turn annotated with \texttt{***}.

\vspace{12pt}

\section{Discourse Analysis on Pseudo-Labeled Data}
\label{app:pseudo_comparison}

This appendix reproduces the key discourse analysis figures using labels
assigned by the best-performing RoBERTa-base (+Aug) model to a sample (i.e.,
ten sessions) unlabeled corpus.  Figures should be compared against their
human-labeled counterparts in the main text.  The aggregated CCI
lesson-progression plot is discussed in Section~\ref{sec:pseudo}; the remaining
figures are presented here.

\begin{figure}[!t]
  \centering
  \includegraphics[width=\linewidth]{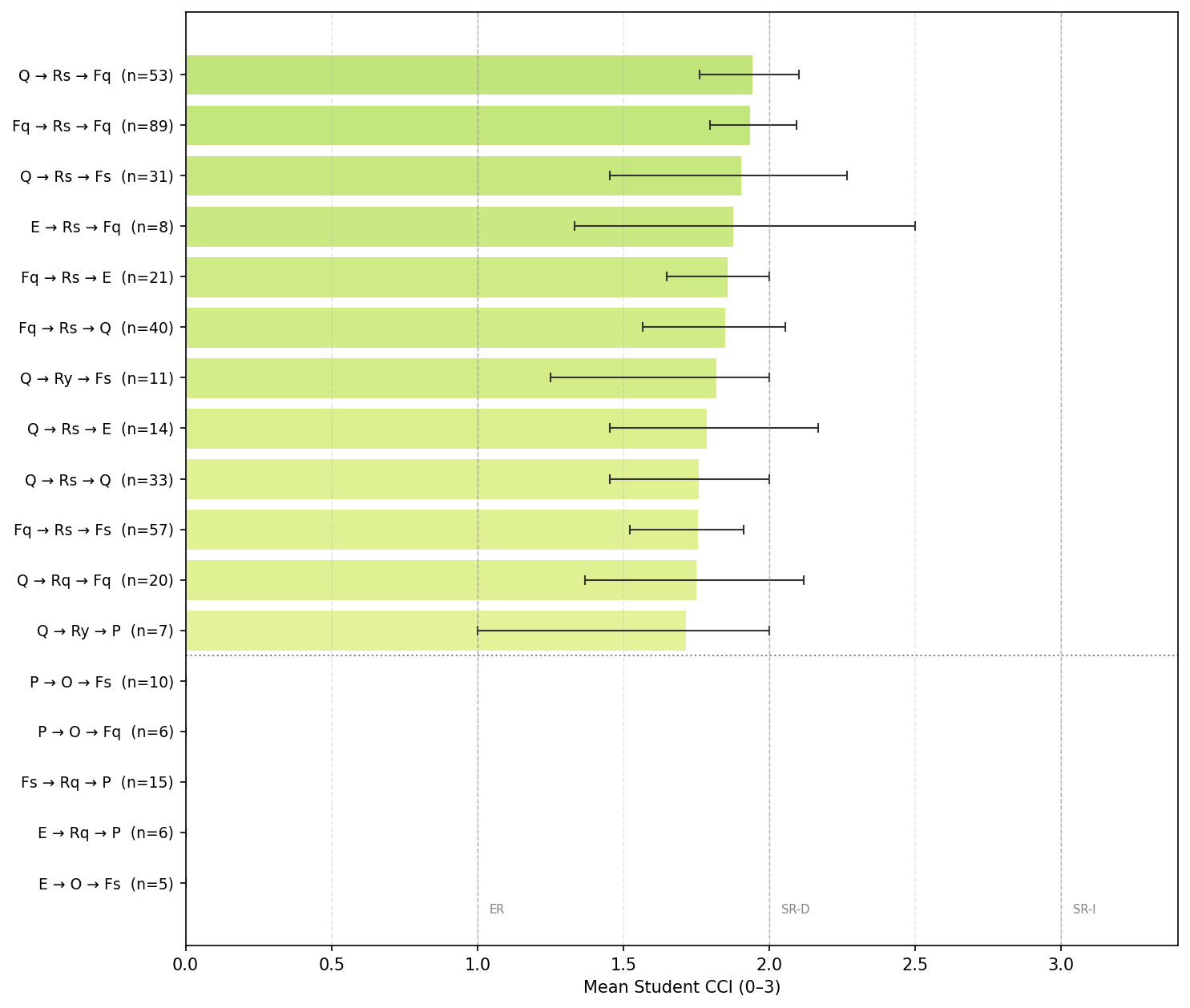}
  \caption{3-turn Teacher--Student--Teacher IRF patterns ranked by mean
    student CCI on pseudo-labeled data (min.\ $n$=5).  Compare with
    Fig.~\ref{fig:3gram_bar} (human-labeled).  Overall CCI values are
    lower and the gap between top- and bottom-ranked patterns is narrower,
    consistent with the model under-predicting higher-order RC classes.}
  \label{fig:3gram_bar_pl}
\end{figure}

\begin{figure}[!t]
  \centering
  \includegraphics[width=\linewidth]{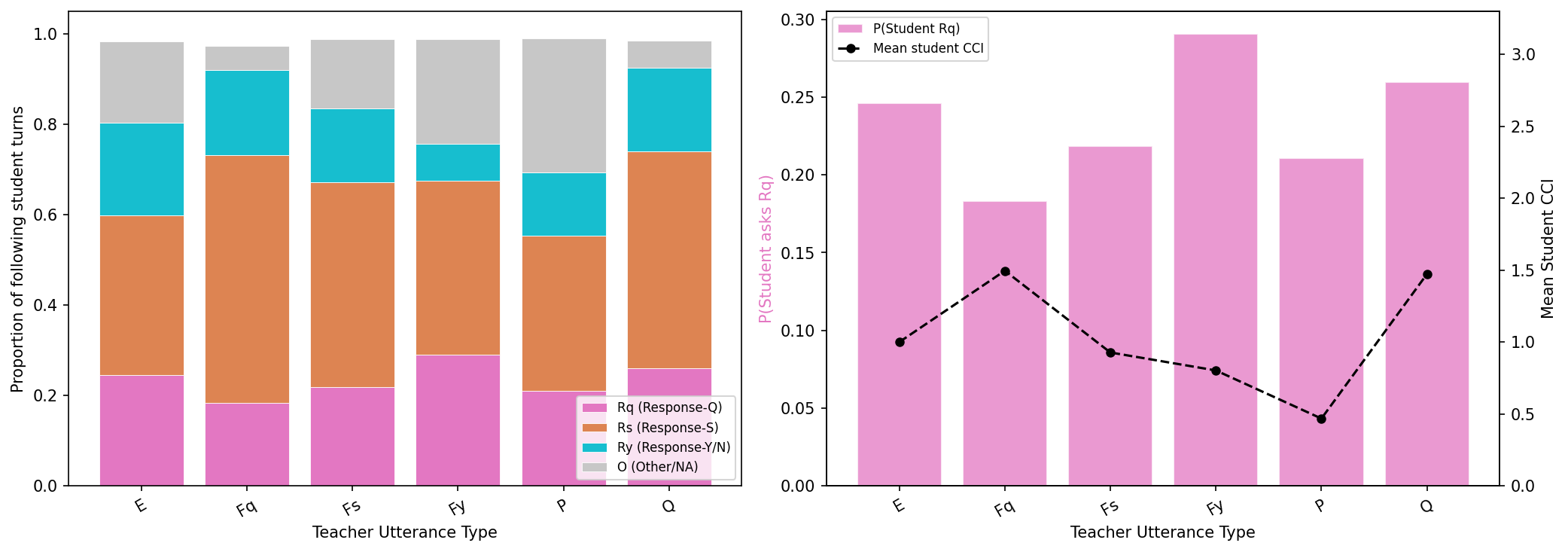}
  \caption{Rq trigger analysis on pseudo-labeled data.  Left:
    $P(\text{student UT} | \text{teacher UT})$ for all lag-1 T$\to$S
    transitions.  Right: $P(\text{student Rq})$ (bars) and mean student
    CCI (line) per preceding teacher UT.  Compare with
    Fig.~\ref{fig:rq_triggers} (human-labeled).}
  \label{fig:rq_triggers_pl}
\end{figure}

\begin{figure}[!t]
  \centering
  \includegraphics[width=\linewidth]{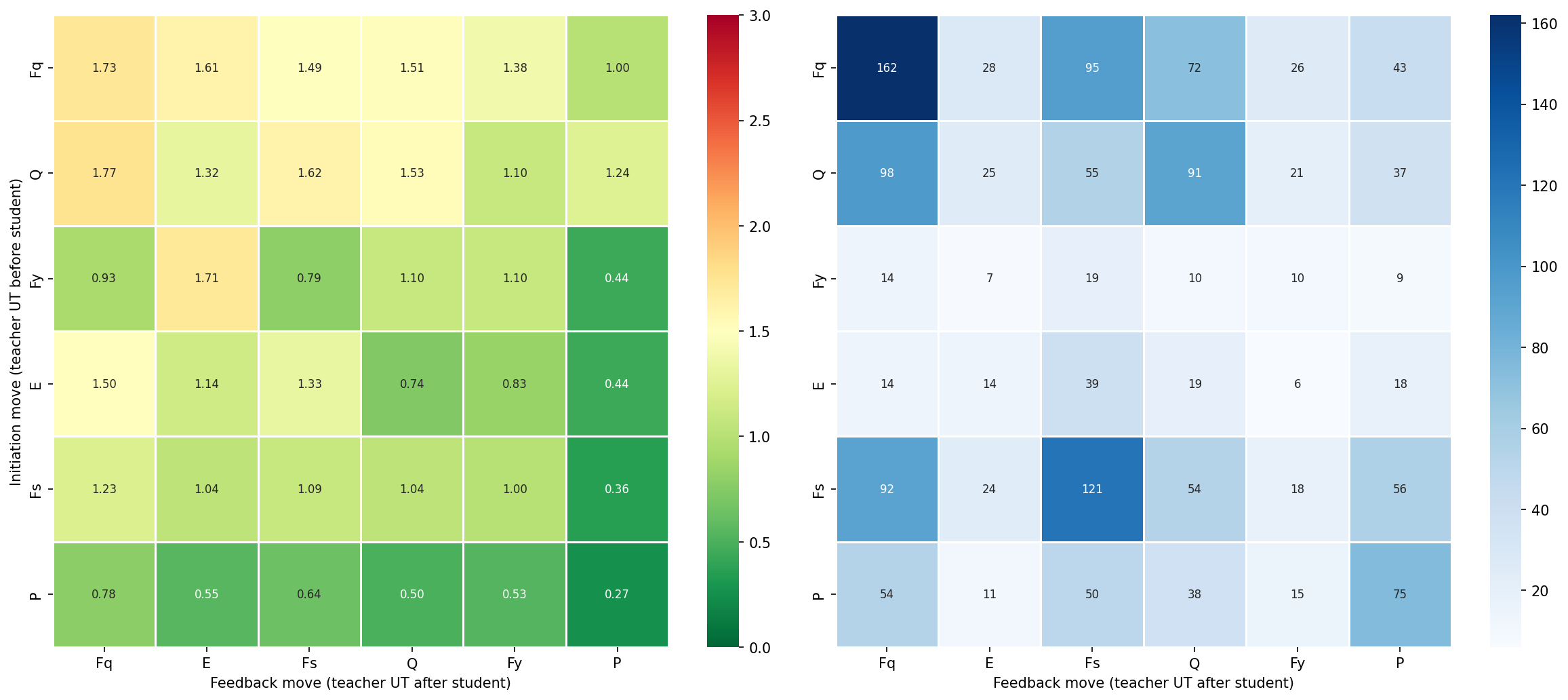}
  \caption{Teacher-framing heatmap on pseudo-labeled data.  Left: mean
    student CCI per (initiation UT $\times$ feedback UT) pair.  Right:
    observation counts.  Compare with Fig.~\ref{fig:framing_heatmap}
    (human-labeled).  CCI values are uniformly lower and the high-CCI
    peak at Fq$\times$Q visible in the human-labeled heatmap is absent,
    reflecting the model's suppression of SR-I predictions.}
  \label{fig:framing_heatmap_pl}
\end{figure}

\end{document}